\pdfoutput=1

\documentclass[11pt]{article}

\usepackage{EACL2023}

\usepackage{tablefootnote}
\usepackage{times}
\usepackage{comment}
\usepackage{latexsym}
\usepackage{multirow}
\usepackage{color}
\usepackage{pgfplots}
\usepackage{ulem}
\usepackage{pifont}

\usepackage[T1]{fontenc}

\usepackage[utf8]{inputenc}

\usepackage{microtype}

\usepackage{inconsolata}

\title{DISTO: Evaluating Textual Distractors for Multi-Choice Questions using Negative Sampling based Approach}

  \author{
Bilal Ghanem$^{*}$ and Alona Fyshe \\
  {\normalsize Department of Computing Science, University of Alberta, Canada} \\
  \small{bilalhgm@gmail.com, alona@ualberta.ca}
}

\begin{document}
\maketitle
\begingroup\def\thefootnote{*}\footnotetext{The current affiliation is TalentNeuron.}\endgroup
\begin{abstract}
Multiple choice questions (MCQs) are an efficient and common way to assess reading comprehension (RC). Every MCQ needs a set of distractor answers that are incorrect, but plausible enough to test student knowledge. Distractor generation (DG) models have been proposed, and their performance is typically evaluated using machine translation (MT) metrics. However, MT metrics often misjudge the suitability of generated distractors. We propose DISTO: the first \emph{learned} evaluation metric for generated distractors. We validate DISTO by showing its scores correlate highly with human ratings of distractor quality.  At the same time, DISTO ranks the performance of state-of-the-art DG models very differently from MT-based metrics, showing that MT metrics should not be used for distractor evaluation.

\end{abstract}

\section{Introduction}
Multiple choice questions (MCQs) are a popular questioning format to assess reading comprehension (RC). Compared to written answers, MCQs allow for quick and automatic evaluation, and consistent scoring. Given a passage, question, and a set of plausible answers, the student needs to select the single correct answer. The main challenge that the student faces in this form of questions is the relatedness of the plausible answers (distractors) to each other, and the semantic context consistency of the plausible answers to the question and context~\cite{goodrich1977distractor}. 

To reduce the effort and time needed to create good distractors for MCQs, models for distractor generation (DG) task have been proposed. However, previous methods to evaluate generated distractors can misjudge performance. Previously, the DG task used machine translation (MT) metrics (e.g BLUE) to evaluate generated distractors~\cite{xie2021diverse,kalpakchi2021bert,rodriguez2022end}. These metrics were not designed to evaluate DG, and so they do not consider important aspects (e.g. semantic relatedness, and context consistency) required to evaluate distractors. To the best of our knowledge, we are the first to propose a specialized learned metric to evaluate the textual distractors in an automatic way.

\begin{figure}

\centering

\includegraphics[width=7.5cm]{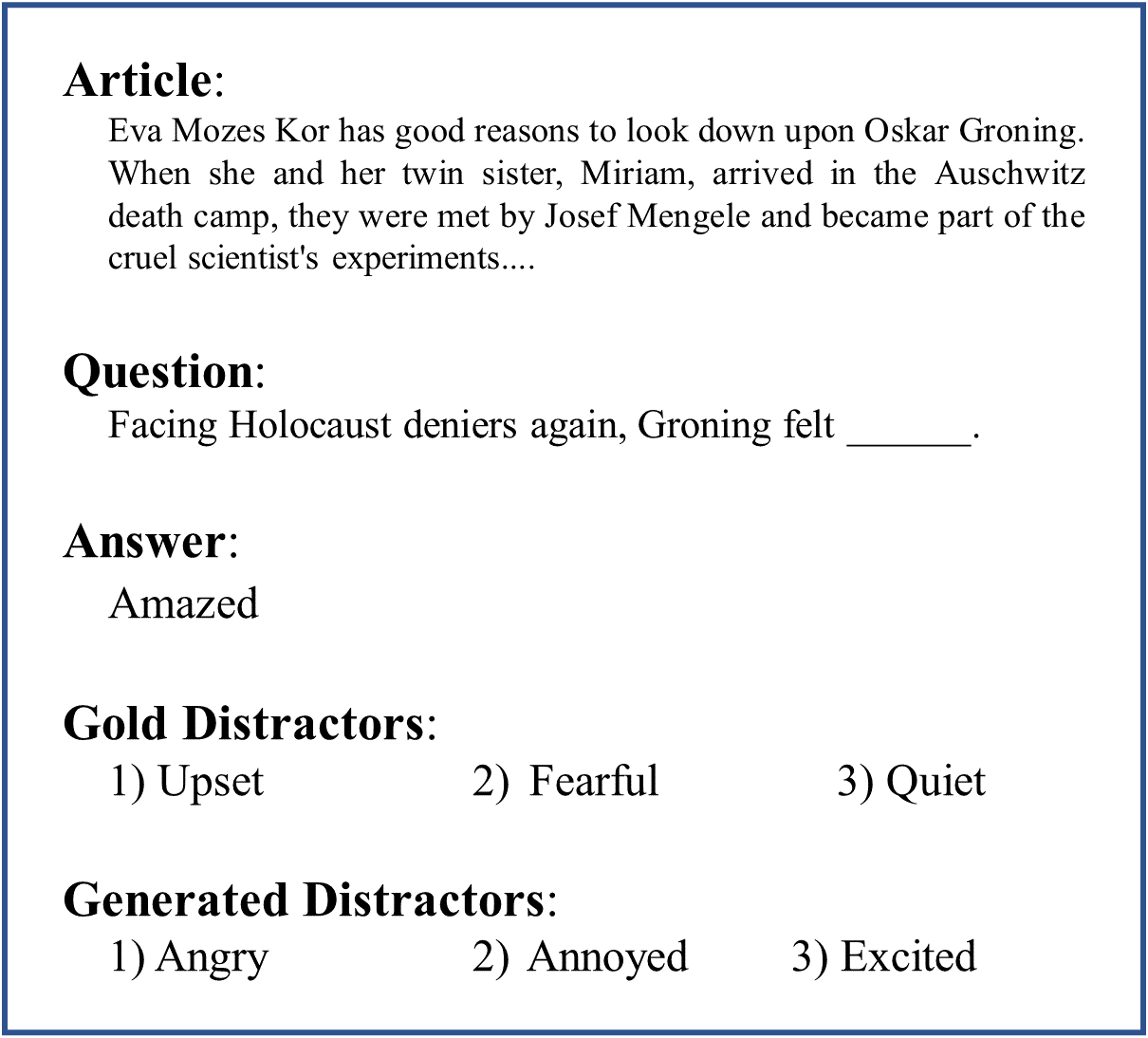}

\caption{A multi-choice question example from the RACE dataset~\cite{lai2017race}. The generated distractors were produced using a T5 model. Though the generated distractors are reasonable, MT metrics would give them a zero score.}

\label{fig:DG_example}

\end{figure}

To clarify the distractor evaluation issue with an example, in Figure~\ref{fig:DG_example} we present an MCQ example from the RACE dataset~\cite{lai2017race} with generated distractors using T5~\cite{raffel2020exploring}. Considering the given context, we can see that the generated distractors are good plausible answers. However, none match the gold distractors. Thus, using MT metrics, such as BLEU~\cite{Papineni2002}, and ROUGE~\cite{lin2004rouge}, these distractors will receive zero scores. Distractor evaluation metrics should not look at the textual overlapping with the gold, as the gold distractors \emph{are not the only possible options}. In Figure ~\ref{fig:DG_example}, we can substitute each of the generated distractors with many other emotions (e.g. happy, sad, grateful, etc.) and the distractors will still be good choices for the question. Additionally, the semantic relatedness of the distractors to the answer is an important aspect. For example, a question that asks about the capital of France should not have ``\textit{Paris Hilton}'' as a distractor since it is not contextually or semantically related to the answer.

In this work, we study the problem of evaluating the generated distractors. For that, we propose the first learned distractor evaluation metric, DISTO, that uses a negative sampling (NS) strategy to learn what a bad distractor looks like. 
We conduct a human evaluation, and show that DISTO's scores correlate highly with human ratings. With evidence of DISTO's reliability, we re-evaluate state-of-the-art DG models using DISTO and find that it produces a different ranking for the performance of those models. Creating learned metrics has been explored in the NLP community for some tasks like MT~\cite{zhang2019bertscore,sellam2020bleurt}.

The contributions in this work are as follows: 

\begin{itemize}
\itemsep0em 
    \item \noindent We introduce a novel problem: distractor evaluation, that has been completely overlooked previously. We show that using MT metrics to evaluate DG models is not suitable.

    \item \noindent We propose DISTO, a distractor evaluation metric that uses a negative sampling technique to measure the consistency of a given set of distractors with respect to the context. Unlike previous approaches, DISTO does not apply any kind of text-based similarities to evaluate the generated distractors~\footnote{The code, data, and trained model are available at: https://github.com/anonymous}.

    \item \noindent We demonstrate the validity of the proposed metric with an extensive evaluation process, including a human evaluation on two sources of data to validate DISTO’s performance.
\end{itemize}

The rest of the paper is structured as follows. In the next section, we present an overview of the literature work for both distractors evaluation and generation. In Section 3, we describe the methodology and the negative sampling technique. The experimental setting, and the results are presented in Section 4. In Section 5, we evaluate the available DG models using DISTO and we compare it to MT metrics. Finally, we draw some conclusions and possible future work for this study.

\section{Related Work}

\begin{table*}[h]
\centering
\small
\begin{tabular}{l|c|c|c|c}
\hline
Study & Approach & Lang. & Domain/Source & Evaluation \\
\hline
\citet{mitkov2003computer}  & Hypernyms from Word-Net lexicon & En & Textbooks & Manual \\
\citet{pino2009semi}        & Phonetic and morphological similarities & En & Pronouncing Dictionary  & Manual \\
\citet{kumar2015revup}      & Word2Vec semantic similarity & En & Textbooks & Manual \\
\citet{guo2016questimator}  & Word2Vec semantic similarity & En & Wiki & N/A~\footnotemark \\
\citet{hill2016automatic}   & Ngrams co-occurrence likelihood & En & Google Ngrams$_\delta$ & Manual \\
\citet{jiang2017distractor} & Word2Vec semantic similarity & Cn & Textbooks, Wiki & Manual \\
\citet{stasaski2017multiple} & Structural similarities in an ontology & En & Educational ontology & Manual \\
\citet{yeung2019difficulty} & BERT with [MASK] filling & Cn & Textbooks & Manual \\
\citet{gao2019generating}   & LSTM encoder-decoder & En & RACE$_\delta$ & MT + Manual \\
\citet{chung2020bert}       & BERT with [MASK] filling & En & RACE$_\delta$ & MT \\
\citet{zhou2020co}          & LSTM encoder-decoder & En & RACE$_\delta$ & MT + Manual \\
\citet{offerijns2020better} & GPT-2 Transformer & En & RACE$_\delta$ & MT + Manual \\
\citet{qiu2020automatic}    & LSTM encoder-decoder & En & RACE$_\delta$ & MT + Manual \\
\citet{maurya2020learning}  & LSTM encoder-decoder(s) & En & RACE$_\delta$ & MT + Manual \\
\citet{kalpakchi2021bert}   & BERT with [MASK] filling & Se & SweQUAD-MC$_\delta$ & Manual \\
\citet{xie2021diverse}      & T5 Transformer & En & RACE$_\delta$, Cosmos QA$_\delta$ & MT + Manual \\
\hline
\end{tabular}
\caption{A survey of the existing distractor generation models. The $_\delta$ sign in the ``Domain/Source'' column means an online published dataset/corpus. For the Language column: En-English, Cn-Chinese, Se-Swedish. MT in the ``Evaluation'' column means Machine Translation metrics.}
\label{tab:related_works}
\end{table*}

\noindent \textbf{Distractors Evaluation.} Having scoring metrics for the DG task is important because the current evaluation techniques are either inaccurate (MT metrics) or very time-consuming (human evaluation). In Table \ref{tab:related_works}, we show past DG techniques, along with methods for evaluates distractor quality. Almost all use manual evaluation which is costly, and cannot give real-time results during experiments. We need a proper automatic metric, but current automatic scoring can produce incorrect results (see example in Figure \ref{fig:DG_example}). In fact, tuning a machine learning model to maximize MT scores could lead to overfitting, which may be one of the reasons DISTO rankings are so different from those previously reported (see Section~\ref{sec:DISTO_FOR_DG_MODELS}).

The related work uses several ways to evaluate the generated distractors manually. Most ask human annotators to assess the quality of distractors using a score range (e.g. from one to five) considering a given set of categories, such as fluency, coherence, relevance, diversity, etc~\cite{kumar2015revup,jiang2017distractor,stasaski2017multiple,yeung2019difficulty,zhou2020co,qiu2020automatic,maurya2020learning,xie2021diverse}. Other works infer those categories after passing students/annotators an MCQ test that includes the new distractors~\cite{mitkov2003computer,hill2016automatic,gao2019generating}. The works ~\cite{offerijns2020better,kalpakchi2021bert} evaluate the quality by asking the annotators questions like: ``\textit{does the answer make sense in relation to the question?}'' or by asking them to select only good distractors. 

\noindent \textbf{Distractor Generation.} Work on DG can be divided mainly into two main lines of research: generative models and ranking models. In the first research line, the models generate distractors for a given MCQ. In the table, we only list generative work as we focus mainly on these approaches for evaluation. In the second research line, the authors frame the task as a ranking problem where a model needs to rank a distractor within a given candidate set~\cite{zesch2014automatic,pho2015distractor,welbl2017crowdsourcing,liang2018distractor,hayaneva2018automatic}. Here, the ranking model assumes that the distractor set exists and it does not need to generate them. This type of models can be directly evaluated using an information retrieval metric since it is a ranking task.

\section{Methodology}
\footnotetext{This work proposed an assessment system using MCQs. The authors focused on evaluating the created questions only.} 

\subsection{Data Augmentation with Negative Sampling}
\label{sec:data_augmentation}
A good DG model needs to generate a set of distractors that are semantically consistent with the context. In order to evaluate that, we propose to learn the distractors' consistency from the currently available RC datasets that were created by human experts. In those datasets, each \textit{Article}-\textit{Question} pair is associated with an answer and $N$ $\in$ [1,3] distractors. Using these datasets, we can only learn what is a good distractor set, but not what make a bad distractor. Thus, we use a NS technique with distractor augmentation to create bad distractors sets. In this way, we can model both cases (good and bad distractors sets) and assign a consistency score for a given distractor in a context.

Given an RC dataset that contains in each instance an article $Ar$, question $Q$, answer $An$, and $N$ distractors (D$_1$..$D_n$), we want to augment those contextually consistent distractors (good distractors) and replace them with inconsistent ones (bad distractors). Our proposed model takes [$Q, An, D, Ar$] as an input and outputs a consistency score [0-1]. 

We design our model to take a \emph{single distractor} in each instance, thus for the instances from the RC datasets where there are more than one distractor, we replicate the instances $N$ times so each contains one of the $N$ distractors with the same [$Q, An, Ar$] set. Since our goal is to design a metric that evaluates DG models, we add a linear layer with a sigmoid function as an output to our model (see Section~\ref{sec:disto_architectures}) to form a regression task. For the instances where we have good distractors, we set the score to one, and for the ones that have augmented distractors (bad distractors), we set it to zero. In this setting, the model outputs 0 for a bad distractor, 1 for an excellent distractor, and a value in between depending on the distractor plausibility. Also, when a DG model generates $N$ distractors, we feed each one with its [$Q, An, Ar$] set and we average the $N$ obtained scores.

In order to build our negative cases, we augment the distractors using one of the following techniques:

\noindent \paragraph{1) Answer Replication:} We duplicate the answer as a distractor to cover the cases when a DG model generates a distractor too similar or identical to the answer.

\noindent \paragraph{2) Random Distractor:} We build a pool of all distractors ($\sim$310K distractors) from the used RC datasets, and then we select one randomly. We make sure that the chosen random distractor does not equal to the current good distractor in the given instance. This technique helps the model to penalize bad generated distractors that are totally inconsistent with the context.

\noindent \paragraph{3) Farthest Point in a Cluster:} The distractors of an MCQ have characteristics in common with the correct answer (similar length, semantically related, etc.). In this technique, we replace a good distractor with a new bad one that shares those characteristics. 
Thus, we use the following set of features for the distractor textual representation:
    \begin{itemize}
    \itemsep0em
    \item \noindent BERT Embeddings to capture the semantic relatedness~\footnote{We use the ``bert-base-uncased'' model from the BERT-as-Service library to extract the embeddings.}.
    
    \item \noindent Bag-of- POS Tags: For each distractor, we build a Term Frequency (TF) vector of POS tags~\footnote{We use the spaCy library to do POS tagging.}. We have noticed that good distractors have similar POS structures. This has been proved in \citet{pho2014multiple}.
    
    \item \noindent Bag-of- Named Entity Types: For each distractor we build a TF vector of named entity types. Similarly, we have noticed that relevant distractors contain similar named entity types.
    
    \item \noindent Distractor Length: Relevant distractors have usually very similar tokens length. Thus, here we count the number of tokens for a given distractor.
    \end{itemize}

After that, using the pool of all the distractors ($\sim$310K distractors) represented with the previous features, we use the K-means clustering algorithm to build clusters of distractors. We set the number of clusters to 200~\footnote{We tested several clusters' numbers [50, 100, 200, 300] and evaluated their results manually. We noticed that 200 gave us the best results.}. After building the clusters, for each good distractor that we want to replace, we choose the farthest point (distractor) in the corresponding cluster using euclidean distance. Our goal is to replace the good distractor with another one that is somewhat close but not too close to fit the context. We noticed that replacing a distractor with the closest distractor in a cluster gives us an augmented distractor that can be considered a good distractor. For instance, for the distractor ``\textit{A tiger named benny}'' the closest distractor in the cluster is ``\textit{A dog called buck}''. On the other hand, the farthest point in the cluster is ``\textit{A midnight madness event}''.

\noindent \paragraph{4) BERT [MASK] Filling:} The initial training task of BERT is to fill masked tokens in tokenized sentences. We leverage this functionality by replacing nouns, verbs, and adjectives in a good distractor to create bad augmented distractor. Every time we apply the mask filling on a token, BERT returns the top N possible tokens with their probabilities to fill the masked token. We discard the associated probabilities and select uniformly from the tokens. We make sure that the chosen token is not the same as the original masked token (before masking). With this technique, we introduce lexical alterations to the good distractor components while maintaining fluency. For instance, the good distractor ``\textit{They focus on bible stories}'' is replaced with ``\textit{They drew on the sand}''.


From each original instance, we create \emph{four negative instances}, each one uses one of the above mentioned augmentation techniques. Each time we substitute a good distractor with an augmented one, we set the score of the new instance to zero. We validated the augmentation techniques manually with a random sample from each technique. For each sample, we had a look at the corresponding article, question, and answer to make sure that they do not fit the context. We found that the augmented distractors are valid bad distractors, where some of them are very far from the given context and others are closer but still invalid. It is worth mentioning that, we found a few cases where the created bad distractors can be considered as good distractors in the given context.

Since our approach is to learn the distractors' consistency from the currently available RC datasets, we use several MCQ datasets. The datasets are CosmosQA~\cite{huang2019cosmos}, DREAM~\cite{sun2019dream}, MCScript~\cite{ostermann2019mcscript2}, MCtest~\cite{richardson2013mctest}, Quail~\cite{rogers2020getting}, RACE~\cite{lai2017race}, and SCIQ~\cite{welbl2017crowdsourcing}. We preprocess these datasets before we apply the NS technique, where we remove instances that have a corrupted answer, question, or article (e.g. empty texts, filled with punctuation marks, etc.), and instances that have ``none of the above'' answer~\footnote{The ``none of the above'' answers will confuse the model because there is no consistency between them and their distractors.}. Also, we remove instances that have no distractors or instances that have redundant distractors. We use the original data splits for all the datasets except Quail where we sample 0.1 for validation as it does not have a defined validation set. In Table~\ref{tab:summary_datasets}, we present the datasets splits' sizes and the final dataset size after applying NS.

\begin{table}[t]
\centering
\begin{tabular}{l|c|c|c}
\hline
Dataset & Train & Val. & Test \\
\hline
Cosmos QA           & 21,397 & 2,726    & 2,369 \\
DREAM               & 6,107  & 2,035    & 2,036 \\
MCScript            & 14,189 & 2,020    & 3,610 \\
MCtest              & 1,200  & 200      & 599 \\
Quail               & 9,215  & 1,025    & 2,164 \\
RACE                & 40,385 & 2,234    & 2,201 \\
SCIQ                & 10,480 & 887      & 883 \\ \hline
Total               & 102,973 & 11,127  & 13,862 \\ \hline
Flattened           & 274,366 & 27,303  & 32,322 \\ \hline
+ NS                & 1,043,464 & 104,485  & 124,724 \\
\hline
\end{tabular}
\caption{A summary of the data splits of the used datasets after the preprocessing step. Flattened: when we duplicate an instance that has $N$ distractors to $N$ instances, each has one of the $N$ distractors.}
\label{tab:summary_datasets}
\end{table}

\subsection{Distractor Evaluation Architectures}
\label{sec:disto_architectures}
To model the consistency of the distractors to the context, we use a pretrained encoder transformer model that takes the [$Q, An, D, Ar$] as an input and outputs a consistency score. We experimented with BERT~\cite{devlin2018bert}, RoBERTa~\cite{liu2019roberta}, Longformer~\cite{beltagy2020longformer}, and the distilled versions of BERT and RoBERTa from HuggingFace library~\cite{wolf2019huggingface}~\footnote{We use the base version of these models.}. We found that the Distilled RoBERTa model gave us the best results in our initial experiments, and we based our experiments on that model. In order to capture the relation between the distractors with the context, we experiment with two architectures:

\noindent \paragraph{Separated Text (SepT):} In this architecture, we feed the input texts separated with special tokens to the encoder model. The input text structure looks like the following: [QUES] $Q$ [ANS] $An$ [DIS] $D$ [ART] $Ar$. After that, we use the first token from the encoder (classification token [CLS]~\footnote{We refer to the classification token in the DistillRoBERTa model as [CLS] although the classification token for this model was renamed to <s>.}) and we feed it to a sigmoid function to map the logits into the 0-1 range.

\begin{figure}[h]
\centering
\includegraphics[width=6.5cm]{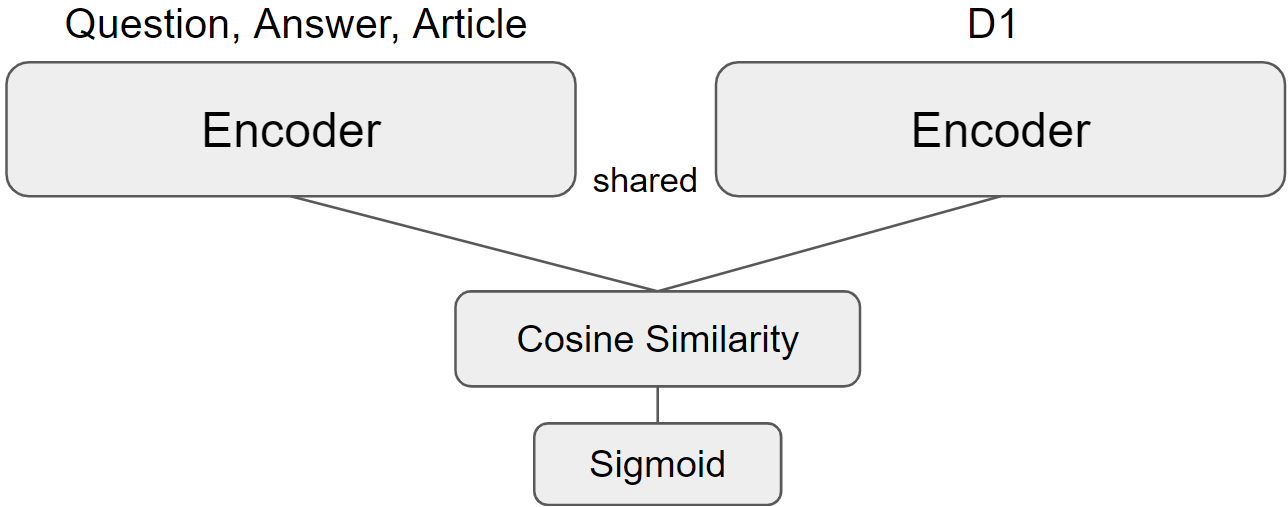}
\caption{The SIAM-COS-SIM model structure.}
\label{fig:model2}
\end{figure}

\noindent \paragraph{SIAM-COS-SIM:} This model uses a Siamese architecture~\cite{mueller2016siamese} with the DistillRoBERTa as an encoder to capture the similarity between the distractors and the context. For that, we feed [QUES] $Q$ [ANS] $An$ [ART] $Ar$ to the first branch and the [DIS] $D$ to the second branch. After that, we measure the cosine similarity between the [CLS] tokens of both branches, and finally we apply sigmoid function of the similarity scores. Our hypothesis in this model is that, if both [CLS] tokens are similar then the distractor is relevant to the context, and thus a good distractor. Figure~\ref{fig:model2} illustrates the structure.

In addition to the two architectures, we create a baseline using Bag-of-words with Tf-Idf weighting scheme and Linear Regression classifier (BOW)~\footnote{We use the a Linear Regression implementation from the Scikit-Learn library.}. Here, we use the same input format as in the SepT architecture ($Q$ [ANS] $An$ [DIS] $D$ [ART] $Ar$) when we transform it to Tf-Idf vectors.

\section{Evaluation of DISTO}

\noindent \textbf{Model Settings and Metrics} For our architectures, we use Adam optimizer~\cite{kingma2014adam} and 1e-5 learning rate value. We set the maximum sequence length to 512. Since we formulate the problem as a regression problem, we use Mean Squared Error (MSE) loss function. In all of our experiments, we use the early stopping regularization technique. To evaluate the models, we use Mean Absolute Error (MAE), and we follow the Workshop on Machine Translation (WMT) Metrics shared task~\cite{mathur2020results} by using the Pearson Correlation~\cite{benesty2009pearson}.

\begin{table}[h]
\centering
\begin{tabular}{l|c|c}
\hline
Model & MAE (\%) & Pearson$_{corr}$ \\
\hline
BOW    & 69.0 & 02.9 \\
SIAM-COS-SIM    & 11.4 & 80.2 \\
SepT (DISTO)    & \textbf{03.8} & \textbf{94.1} \\
\hline
\end{tabular}
\caption{Evaluation results of the distractors evaluation models.}
\label{tab:disto_results}
\end{table}


\subsection{Results and Human Evaluation}
\label{results_human_evaluation}
Table~\ref{tab:disto_results} presents the evaluation results. The results show that there is a huge gap between both of the proposed architectures compared to the baseline. We attribute this due to the ability of both architectures to model the semantics of the inputs, where the Tf-Idf vectors in the BOW are not able to get the semantic meaning. This actually confirms the importance of the semantic meaning for this task. Regarding the two architectures, SepT achieves the highest results with a very high positive correlation and almost zero MAE value. This is probably because one of the training objectives of the transformers-based encoder models is the Next Sentence Prediction (NSP) given the former one. Our setting in the SepT model is very similar to that training objective, and this does not make the model update excessively the encoder weights to learn a new task. In addition to that, the NSP task has in common an important aspect with our task which is semantic relatedness. In the rest of our experiments, we use SepT model and we refer to it as DISTO.

To validate the performance of the DISTO model, we conduct a human evaluation experiment using Amazon Mechanical Turk. For this experiment, we use 50 original instances from the pool of the all datasets' instances where the distractors are good and another 50 instances from the NS technique where the distractors are bad (in total 100 instances). We use instances from the NS technique that contain distractors created only using either the ``Farthest Point in a Cluster'' or ``BERT [MASK] Filling'' techniques, as the bad distractors that were created by the other techniques are easily spotted (duplicated answer and random -based distractors). In this experiment, we show the workers one distractor with its article, question, and answer. We ask the workers to rate the distractor as either bad, neutral, or good, within the given context. In Appendix~\ref{sec:appendix:AMT}, we show a sample from the annotation interface. Because this is announced task, we conducted a manually prepared short quiz (see Figure \ref{fig:AMT_sample}) for the workers to select the strongest. From our final list of good workers, we employ five annotators to annotate each given instance. For the final score from the workers, we average their ratings. Since this is a relatively hard task, we add instructions and a couple of examples to the workers to make sure that they understand completely the task.

\begin{table}[h]
\centering
\begin{tabular}{l|c|c}
\hline
Experiment              & Pearson$_{corr}$ & P$_{value}$\\
\hline
Gold Data vs. Workers   & 0.78 & < 0.001 \\
Gold Data vs. DISTO     & 0.94 & < 0.001 \\
Workers vs. DISTO       & 0.81 & < 0.001 \\
\hline
\end{tabular}
\caption{Pearson Correlation between DISTO, the AMT workers averaged ratings, and the gold data. P$_{value}$ ranges: $\le$~0.001, $\le$~0.05, >~0.05 (Not significant).}
\label{tab:human_corelation_results_on_disto_data}
\end{table}

After we ran the task, we found that one of the five workers was performing poorly in the task (less than 30\% accuracy according to the gold labels). Thus, we discarded this worker and focused on the remaining four workers. We compute the annotation agreement among the workers, and we find a moderate agreement with a Fleiss-Kappa~\cite{fleiss1971measuring} value of 0.45. This shows the difficulty of the task; even for humans it is hard to agree on what is a good or bad distractor. Also, we can conclude from this result that the ``Farthest Point in a Cluster'' and ``BERT [MASK] Filling'' techniques produce distractors that are not easily discarded by the annotators. After that, we measure the Pearson Correlation between DISTO, the workers averaged ratings, and the gold data. Table~\ref{tab:human_corelation_results_on_disto_data} presents the results.

We can see that the human annotations (Workers) correlate highly with the gold data but lower than how DISTO does with the gold data (Pearson Correlation 0.78 vs. 0.94). This demonstrates the difficulty of the task for the human annotators. Also, the results demonstrated the effectiveness of DISTO since it was trained on a large set of data created by professional educators. Finally, the table shows that DISTO correlates with the human annotations with a Pearson Correlation value of 0.81. In general, all of the correlation results are high and significant, especially for the ``Gold Data vs. DISTO'', as we noticed with the complete test set (see Table~\ref{tab:disto_results}).

As we show in the example in Figure 1, MT metrics are not valid evaluators for DG since they consider the gold distractors as the only correct ones. By looking at the above results, we can conclude that DISTO is a coherent evaluation metric for DG that utilizes the semantics of the distractors to evaluate their validity within a given context.

\subsection{Context Importance}

In order to model the consistency of the distractors with the context, we use the article, question, and answer inputs to represent the context. In this experiment, we run an ablation test on the context to examine the importance of each input for the final decision in DISTO. Thus, we train DISTO without considering one of the inputs and then we compare the new result to the original where the whole context is considered. Table~\ref{tab:disto_ablation_results} presents the results, which show that including the answer in the context is the most valuable to the model, as removing it results in a 14.2\% and 22.9\% drop in terms of MAE and Pearson Correlation, respectively. The results also show, unexpectedly, that including the question in the context is the least important. This leads us to conclude that the semantic relatedness between distractors and the answer is most important, where distractors and questions tend to have a weaker semantic correlation. It is also important to include the articles because the tokens/phrases can mean different things in different texts (contexts) which helps to determine the semantics of the distractors.

\begin{table}[h]
\centering
\begin{tabular}{l|c|c}
\hline
Model & MAE (\%) & Pearson$_{corr}$ \\
\hline
DISTO            & 03.8 & 94.1\\ \hline
DISTO - Question & 05.5 & 91.1 \\
DISTO - Article  & 08.1 & 88.1 \\
DISTO - Answer   & 18.0 & 71.2 \\
\hline
\end{tabular}
\caption{Evaluation results of the distractor evaluation models. ``DISTO - X'' means without X included in the context.}
\label{tab:disto_ablation_results}
\end{table}


\section{DISTO for DG Models}
\label{sec:DISTO_FOR_DG_MODELS}
Several DG models have been proposed in the literature, using different training algorithms. As we can notice in Table~\ref{tab:related_works}, transformer models are a popular approach to generating distractors in the last few years. In this section, we re-evaluate those models using our DISTO model and we compare the results to the MT metrics that have been used in previous work.

In this experiment, we use two models from the literature work, which are \citet{gao2019generating}, and \citet{chung2020bert}.~\footnote{To the best of our knowledge, these models are the only available ones online.} In addition, we use another two transformers models: T5~\cite{raffel2020exploring} and GPT-2~\cite{radford2019language} -based models. Those two models were used in \citet{offerijns2020better}, and \citet{xie2021diverse}. We use the RACE dataset~\cite{lai2017race} for evaluation. The authors of \citet{gao2019generating} and \citet{chung2020bert} models use an edited version of the RACE dataset. Thus, in this experiment, we train all the models on the original version of the RACE dataset for a fair comparison. Since each of these models generates three distractors for a given context, we feed the context three times, every time with one of the three distractors and then we average the scores.
We experiment with the following models:

    \noindent \paragraph{1) GDRCQ:} This model ~\cite{gao2019generating} uses an LSTM~\cite{hochreiter1997long} encoder-decoder model with dynamic and static attentions. This method is an example of distractor generation models based on RNN seq2seq architectures. The authors use Glove word embeddings~\cite{pennington2014glove} for initialization and finetune them during the training process.
    
    \noindent \paragraph{2) BDG:} This approach uses the initial training task from BERT (filling masked tokens) to generate distractors~\cite{chung2020bert}. Given the context, the approach appends a [MASK] token at the end of the context to let the model generate the distractors, word by word. Also, the work proposes an answer-negative regularization technique to overcome the answer duplication problem.
    
    \noindent \paragraph{3) GPT-2:} This model uses the GPT-2~\cite{radford2019language} transformer model to generate distractors. We follow the instructions from~\citet{offerijns2020better} to implement the model. For decoding, we use beam search sampling with a beam size of 6. Also, we use Adam optimizer with a 3e-4 learning rate.
    
    \noindent \paragraph{4) T5:} The T5 language model~\footnote{We use the T5-base version from the Huggingface library.} achieved SOTA results on several generative NLP tasks, so we employ it here to generate distractors. Here, we propose a T5 model that takes the article, question, and the answer, and generates three distractors at once with a separator token [SEP] between them. For decoding, we use Nucleus sampling (Top-p)~\cite{holtzman2019curious} with a 0.9 P value~\footnote{We tested Beam Search as well, but we found that Top-P gives better results for this model.}. Also, we use Adam optimizer with a 5e-4 learning rate.
    
    \noindent \paragraph{5) T5$_{disjoint}$:} Natural language generation tasks can utilize multiple correct references at once, as in MT~\cite{zheng2018multi}, image captioning~\cite{karpathy2015deep}, and question generation~\cite{jiaetal2020ask} tasks. This T5 model is trained to generate one distractor at a time. Then, we apply a min-loss function following~\citet{jiaetal2020ask} to generate several diverse distractors. During generation, we use the Diverse Beam Search sampling method~\cite{vijayakumar2016diverse} to generate three distractors for each input. Similar to the T5 model, we use Adam optimizer with a 5e-4 learning rate.


Once we have these models trained, we evaluate them using the BLEU MT metric. Additionally, we use the BLEURT~\cite{sellam2020bleurt} learned metric which is a SOTA MT metric. In Table~\ref{tab:bleu_results}, under the ``MT Evaluation'' header, we present BLEU 1-4 and BLEURT results on the RACE test set. We can see that the BDG model clearly outperforms the other models considering all BLEU lengths, except for BLEU-1, followed by the T5$_{disjoint}$ model. Similarly, for BLEURT, the BDG model performs best, followed by the GPT-2 model.

\begin{table*}[h]
\centering
\small
\begin{tabular}{l|c|c|c|c|c||c|c|c}
\hline
\multirow{2}{*}{Model} & \multicolumn{5}{c||}{\textbf{MT Evaluation}} & \multicolumn{3}{c}{\textbf{Distractor Evaluation}} \\
\cline{2-9}
 & {BLEU-1} & { BLEU-2} & { BLEU-3} & { BLEU-4} & { BLEURT} & { DISTO} & { AnsDup} & { DisDup} \\
\hline
GDRCQ           & 19.0 & 05.3 & 01.6 & 00.6                                        & 22.97 & \underline{82.47} & \underline{0.90} & \underline{1.00} \\
BDG             & \underline{30.2} & \textbf{18.9} & \textbf{13.0} & \textbf{08.9} & \textbf{31.90} & 67.25 & \underline{0.90} & 6.70 \\
GPT-2           & 19.9 & 03.9 & 00.9 & 00.3                                        & \underline{31.07} & 68.75 & \textbf{0.20} & \textbf{0.10}\\
T5              & 25.1 & 08.4 & 02.7 & 00.9                            & 30.10 & 81.75 & \textbf{0.20} & 3.90 \\
T5$_{disjoint}$ & \textbf{32.0} & \underline{13.7} & \underline{05.6} & \underline{02.3}    & 26.42 & \textbf{92.91} & 2.60 & 6.10 \\
\hline
\end{tabular}
\caption{Evaluation of the distractor generation models using the BLEU, BLEURT, DISTO, Answer Duplication (AnsDup), and the Distractor Duplication (DisDup) metrics. Lower is better for AnsDup and DisDup.}
\label{tab:bleu_results}
\end{table*}

We evaluate the models using DISTO, and we include two additional metrics a) \textit{Answer Duplication (AnsDup)}: the percentage of answer-distractor pairs that are an exact match within an instance; b) \textit{Distractor Duplication (DisDup)}: the percentage of distractor-distractor pairs that are an exact match within an instance. The two metrics have scores ranging from 0 to 100, and for both of them, lower is better. Using these two metrics, we can get better insight into the models' performances.

In Table~\ref{tab:bleu_results}, under ``Distractor Evaluation'', we present the DG models' results using DISTO, AnsDup, and DisDup metrics. Notice that DISTO gives a completely different ranking compared to the MT results. The model that performs the best in terms of BLEU and BLEURT metrics (BDG) has the lowest DISTO result, and the second best BLEURT model (GPT-2) performs the second worst model considering DISTO. Overall, the T5$_{disjoint}$ model performs the best. GDRCQ and T5 models have competitive performance with 0.72\% DISTO difference.
By looking at the \textit{AnsDup} and \textit{DisDup} columns, we can see that the values (percentages) are very low and close to each other. We can conclude that none of the models highly suffer of a duplication issue. We can see that the best model, T5$_{disjoint}$, has the highest number of duplications with respect to the answer (0.2\%), and the highest \textit{DisDup} result (6.10\%). This shows that although this model generates the most relevant distractors given the context (highest DISTO value), these distractors sometimes overlap with each other or with the answer. Interestingly, we can see that the GPT-2 model has the lowest \textit{DisDup} value. We manually checked the output of the model and found that the model in some cases makes small changes to the second and third generated distractors. Those small changes prevent the exact matching \textit{DisDup} metric from counting them as redundant cases.
By conducting this evaluation, we can clearly see that MT metrics are providing misleading conclusions regarding DG models. Indeed, the MT metrics assign the highest score to a model that generates distractors that are not the most contextually consistent. On the other hand, a model that generates relevant distractors appears weak by MT metrics.

\begin{table}[h]
\centering
\begin{tabular}{l|c|c}
\hline
Model           & Pearson$_{corr}$ & P$_{value}$\\
\hline
GDRCQ           & 0.75 & $\le$~0.001 \\
BDG             & 0.3  & $\le$~0.05  \\ 
GPT-2           & 0.28 & $\le$~0.05  \\ 
T5              & 0.63 & $\le$~0.001 \\ 
T5$_{disjoint}$ & 0.6  & $\le$~0.001 \\ 
\hline
\end{tabular}
\caption{Co-relation between DISTO and AMT workers based on the generated distractors of the DG models. P$_{value}$ ranges: $\le$~0.001, $\le$~0.05, >~0.05 (Not significant).}
\label{tab:human_corelation_results}
\end{table}


\subsection{Out-of-Domain Human Evaluation}
In Section~\ref{results_human_evaluation}, we validated DISTO model using a human evaluation experiment on a sampled data from the test set of the augmented data that we built for DISTO. This makes that experiment domain-dependent, as DISTO is trained on the same type of the augmented data (might be biased towards the data). Instead in this section, we conduct another human evaluation using the generated distractors of the DG models from the previous section. In this way, we make sure that DISTO does not see distractors created by any of the augmentation techniques in Section~\ref{sec:data_augmentation}. Similar to our previous experiment, we sample 100 instances from their outputs and make sure that the 100 instances from each model corresponds to the same question, answer, and article set. Also, we follow the same settings regarding the AMT workers (number of workers, quiz, etc.).

The results in Table~\ref{tab:human_corelation_results} show that DISTO has a varied correlation with the human annotators with statically significant association. We argue that this varied correlation is due to 1) the weak performance of the human workers and 2) out-of-domain performance's drop in DISTO. We noticed from the workers' answers to the quiz that it is hard for them to decide what is a good or a bad distractor, also, they had very few responses where all the workers' answers are the same. Actually, this was proved in Table~\ref{tab:human_corelation_results_on_disto_data} where ``Gold Data vs. Workers'' correlation value is lower than ``Gold Data vs. DISTO''. Although we are using out-of-domain data that is generated by several DG models which vary in performance, DISTO and the human annotators still show significant correlation with each other. This demonstrates that DISTO is unbiased and domain-independent.

\section{Conclusion and Future Work}
The existence of distractors evaluation metrics is important for the proper evaluation of the new DG models. This importance is not limited to the final evaluation process, instead, it's essential to have an automated metric during the experimentation process to monitor the amount of improvement at each stage and to allow researchers to pick the best model after a hyper-parameter searching process. An incorrect or imprecise evaluation process may lead to invalid results or selection of a weak model. In this work, we studied the problem of textual distractor evaluation. We proposed DISTO which is the first distractor evaluation metric. Unlike MT metrics that use a text-based comparison process to compare generated distractors to the gold ones, DISTO uses a negative sampling strategy with distractors augmentation techniques to model the consistency of distractors within a given context. We validated DISTO with extensive experiments coupled with a human evaluation. 
Our results show that DISTO is accurate and correlates highly with human ratings. Previous evaluation using MT metrics may be invalid and would lead to incorrect conclusions.
In the future we plan to extend our work in two directions: First, we plan to integrate more augmentation techniques to cover more negative cases by including, e.g., grammatical modifications.
Second, we will work to make DISTO multilingual to support the evaluation of distractors in other languages. 
We hope that DISTO and our exploration of distractor evaluation fosters new conversations in the DG community.

\section*{Limitations}
In this work, we integrated several augmentation techniques that can help us to generate bad distractors. However, we assume that the current DG models generate grammatically correct distractors, thus we do not create augmented instances that cover grammatically incorrect distractors. We are not sure how DISTO will handle those cases in its current form.

Using both ``Farthest Point in a Cluster'' or ``BERT [MASK] Filling'' distractors augmentation techniques, we were able to create new bad distractors that are lexically modified. We found that these techniques are very effective to modify the original distractors in a way that the new distractors share some characteristics with the original ones, but at the same time, they are sometimes less contextually relevant. However, this was not always the case. We found in some cases that the two aforementioned techniques generate new good distractors. These good distractors might confuse our model since we assign low scores for them but they are contextually consistent. Finally, we want to highlight that those two techniques are computationally expensive, especially the ``Farthest Point in a Cluster'' technique.

\section*{Ethics Statement}
\noindent \textbf{Human Annotation.} We estimated the amount of time AMT workers need to finish a HIT and then we compensated them so that the payment rate was higher than the local living wage per hour. Each AMT worker received \$0.4 USD for completing one HIT, which we estimated would take on average less than one minute.

\noindent \textbf{Bias in Language Models.} Language models have several types of bias, e.g. gender, race, religion, etc., and this is due to the data used to train them \cite{liang2021towards}. 
We acknowledge that the DISTO model we trained might cause ethical concerns, e.g. assigning a high score to biased distractors.


\bibliography{custom}
\bibliographystyle{acl_natbib}

\appendix
\section{Amazon Mechanical Turk (AMT) Annotation Interface}
\label{sec:appendix:AMT}
\begin{figure*}[h]
    \centering
    \includegraphics[width=16cm]{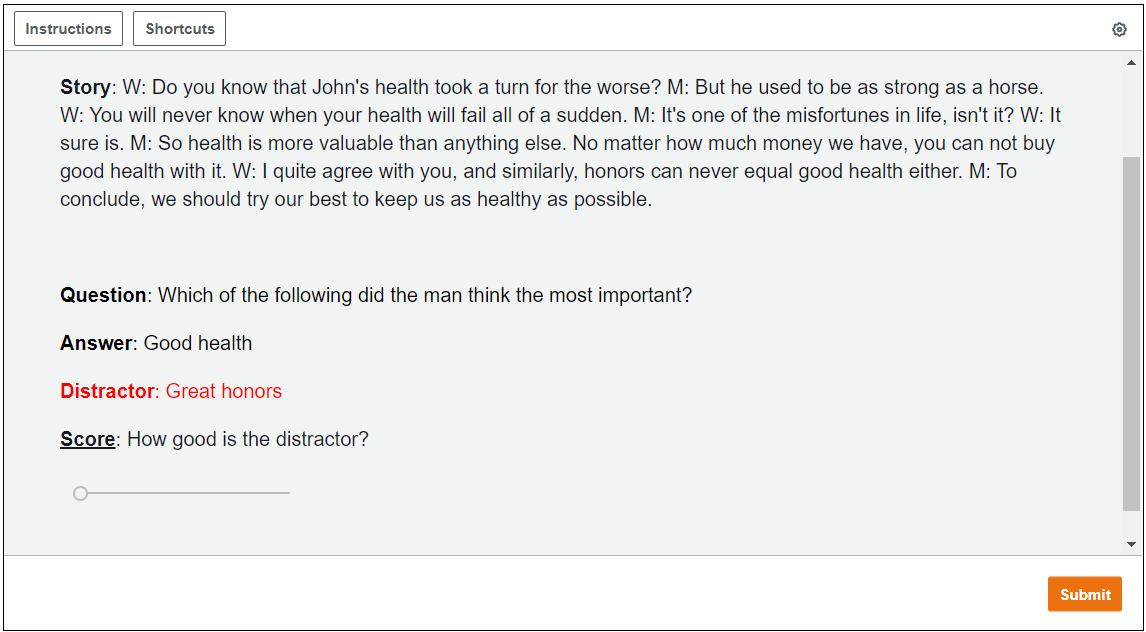}
    \caption{A sample from the AMT interface.}
    \label{fig:AMT_sample}
\end{figure*}

In Figure~\ref{fig:AMT_sample} we present a sample from the AMT interface.

\end{document}